\documentclass[12pt]{article}
\usepackage[pdftex]{graphicx}
\usepackage{amsbsy,amssymb,latexsym,amsmath}
\usepackage{wrapfig}
\usepackage{algorithm}
\usepackage{algpseudocode}
\usepackage{multirow}
\usepackage{amsmath,amssymb,amsfonts}
\usepackage{fullpage}
\usepackage{subfigure}
\usepackage{graphicx}
\usepackage{amsmath}
\usepackage{xfrac}
\usepackage{mathtools}
\usepackage{mathrsfs}
\providecommand{\keywords}[1]{\textbf{\textit{Index terms---}} #1}
\usepackage{url}

\usepackage{soul}
\usepackage{color}
\definecolor{limegreen}{rgb}{0.2, 0.8, 0.2}
\definecolor{forestgreen}{rgb}{0.13, 0.55, 0.13}
\definecolor{greenhtml}{rgb}{0.0, 0.5, 0.0}
\newcommand{\captionsize}{\fontsize{7.97pt}{1.18em}\selectfont}
\newtheorem{example}{Example}
\newtheorem{definition}{Definition}

\include{definitions}

\def\*tr{^{\,*T}}

\title{Distributed Impedance Control of Latency-Prone Robotic Systems with Series Elastic Actuation}
\author{Ye Zhao$^1$ and Luis Sentis$^2$
\thanks{$^1$The author is with The George W. Woodruff of Mechanical Engineering, Georgia Tech, USA, and the Department of Mechanical Engineering, UT Austin, USA. $^2$The author is with the Department of Aerospace Engineering and Engineering Mechanics at UT Austin, USA (corresponding email: ye.zhao@me.gatech.edu)}
}

\date{\vspace{-5ex}}

\begin{document}

\maketitle

\begin{abstract}

Robotic systems are increasingly relying on
distributed feedback controllers to tackle complex and latency-prone sensing
and decision problems. These demands come
at the cost of a growing computational burden and, as a result, larger controller latencies. To maximize robustness to mechanical disturbances and achieve high control performance, we emphasize the necessity for executing damping feedback in close proximity to the control plant while allocating stiffness feedback in a latency-prone centralized control process. 
Additionally, series elastic actuators (SEAs) are becoming prevalent in torque-controlled robots during recent years to achieve compliant interactions with environments and humans. However, designing optimal impedance controllers and characterizing impedance performance for SEAs with time delays and filtering are still under-explored problems. The presented study addresses the optimal controller design problem by devising a critically-damped gain design method for a class of SEA cascaded control architectures, which is composed of outer-impedance and inner-torque feedback loops. Via the proposed controller design criterion, we adopt frequency-domain methods to thoroughly analyze the effects of time delays, filtering and load inertia on SEA impedance performance. These
results are further validated through the analysis, simulation, and experimental testing on high-performance actuators and on an omnidirectional mobile base.


\end{abstract}
\keywords{Robotics, Distributed impedance control, Time delays, Series elastic actuator.}


\section{Introduction}

\begin{figure}[t]
 \centering
   \includegraphics[width=0.5\linewidth]{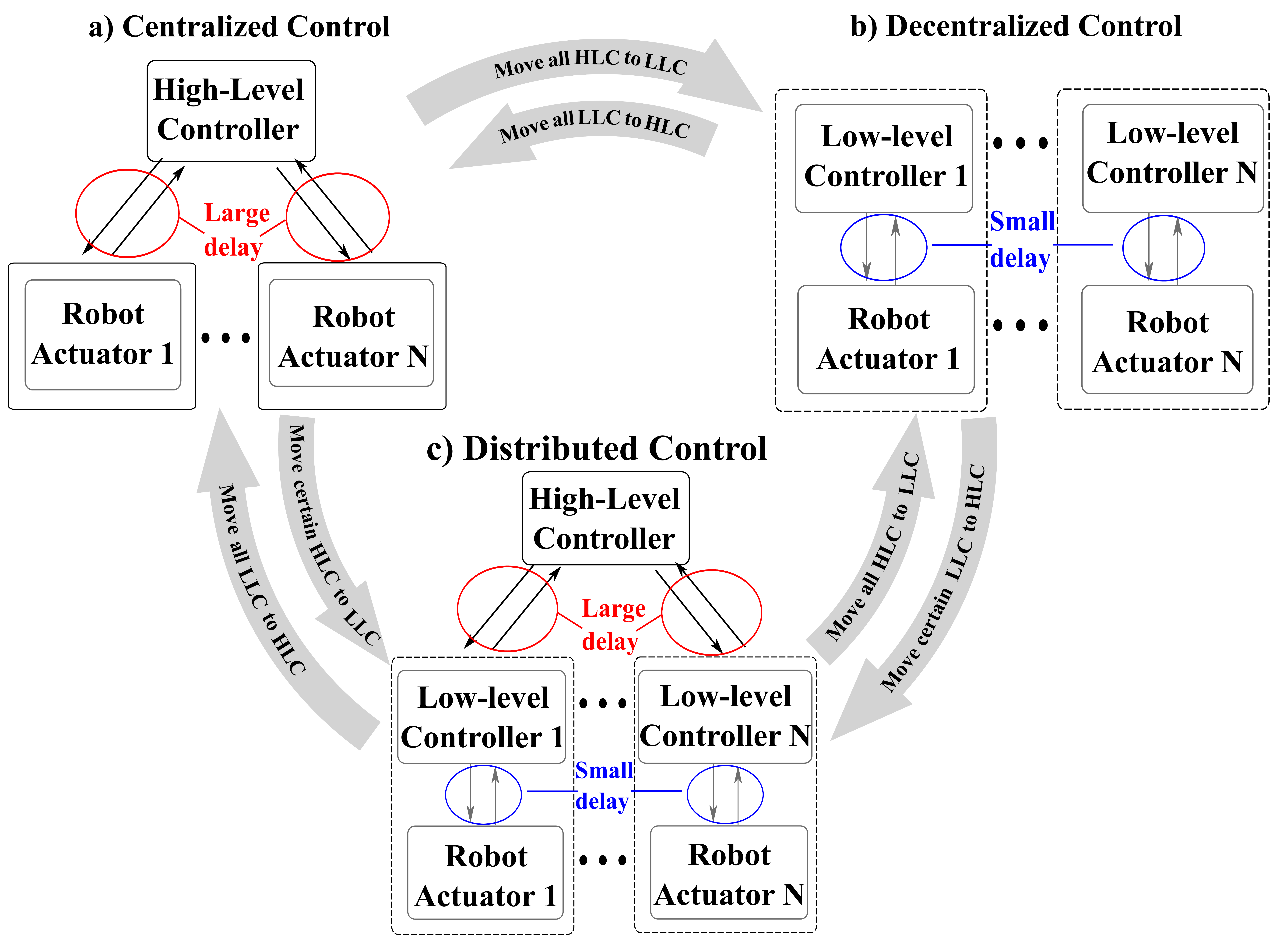}
 \caption{\captionsize Depiction of various control architectures. Many control systems today employ one of the control architectures above: a) Centralized control with only high-level feedback controllers (HLCs); b) Decentralized control with only low-level feedback controllers (LLCs); c) Distributed control with both HLCs and LLCs, which is the focus of this chapter.}
 \label{fig:model}
\end{figure}

As a result of the increasing complexity of robotic control systems, such as human-centered robots \cite{sakagami2002intelligent, diftler2011robonaut} and industrial surgical machines \cite{okamura2004methods}, new system architectures, especially distributed control architectures \cite{kim2005system, santos2006design}, are often being sought for communicating with and controlling the numerous device subsystems. Often, these distributed control architectures manifest themselves in a hierarchical control fashion where a centralized controller can delegate tasks to subordinate local controllers (Figure \ref{fig:model}). As it is known, communication between actuators and their low-level controllers can occur at high rates while communication between low- and high-level controllers occurs more slowly. The latter is further slowed down by the fact that centralized controllers tend to implement larger computational operations, for instance to compute system models or coordinate transformations online.

One concern is that feedback controllers with large delays \cite{karimi2010new, gao2016bounds}, such as the centralized controllers mentioned above, are less stable than those with small delays, such as locally embedded controllers. Without the fast servo rates of embedded controllers, the gains in centralized controllers can only be raised to limited values, decreasing their robustness to external disturbances \cite{lu2014performance} and unmodelled dynamics \cite{martin1981continuous}.

As such, why not remove centralized controllers altogether and implement all feedback processes at the low-level? Such operation might not always be possible. For instance, consider controlling the behavior of human-centered robots (i.e. highly articulated robots that interact with humans). Normally this operation is achieved by specifying the goals of some task frames such as the end effector coordinates. One established option is to create impedance controllers on those frames and transform the resulting control references to actuator commands via operational space transformations \cite{Khatib:87(2)}. Such a strategy requires the implementation of a centralized feedback controller which can utilize global sensing data, access the state of the entire system model, and compute the necessary models and transformations for control. Because of the aforementioned larger delays on high-level controllers, does this imply that high gain control cannot be achieved in human-centered robot controllers due to stability problems? It will be shown that this may not need to be the case. But for now, this delay issue is one of the reasons why various currently existing human-centered robots cannot achieve the same level of control accuracy that it is found in high performance industrial manipulators. More concretely, this study proposes a distributed impedance controller where only proportional (i.e., stiffness) position feedback is implemented in the high-level control process with slow servo updates. This process will experience the long latencies found in many modern centralized controllers of complex human-centered robots. At the same time, it contains global information of the model and the external sensors that can be used for operational space control. For stability reasons, our study proposes to implement the derivative (i.e., damping) position feedback part of the controller in low-level embedded actuator processes which can therefore achieve the desired high update rates.

As it will be empirically demonstrated, the benefit of the proposed split control approach over a monolithic controller implemented at the high-level is to increase control stability due to the reduced damping feedback delay. As a direct result, closed-loop actuator impedance may be increased beyond the levels possible with a monolithic high-level impedance controller. This conclusion may be leveraged on many practical systems to improve disturbance rejection by increasing gains without compromising overall controller stability. As such, these findings are expected to be immediately useful on many complex human-centered robotic systems.

To demonstrate the effectiveness of the proposed methods, this study implements tests on a high performance actuator followed by experiments on a mobile base. First, a position step response is tested on an actuator under various combinations of stiffness and damping feedback delays. The experimental results show high correlation to their corresponding simulation results. Second, the proposed distributed controller are applied to an implementation into an omnidirectional base. The results show a substantial increase in closed-loop impedance capabilities, which results in higher tracking accuracy with respect to the monolithic centralized controller counterpart approach.

\begin{figure}[t]
 \centering
   \includegraphics[width=.7\linewidth]{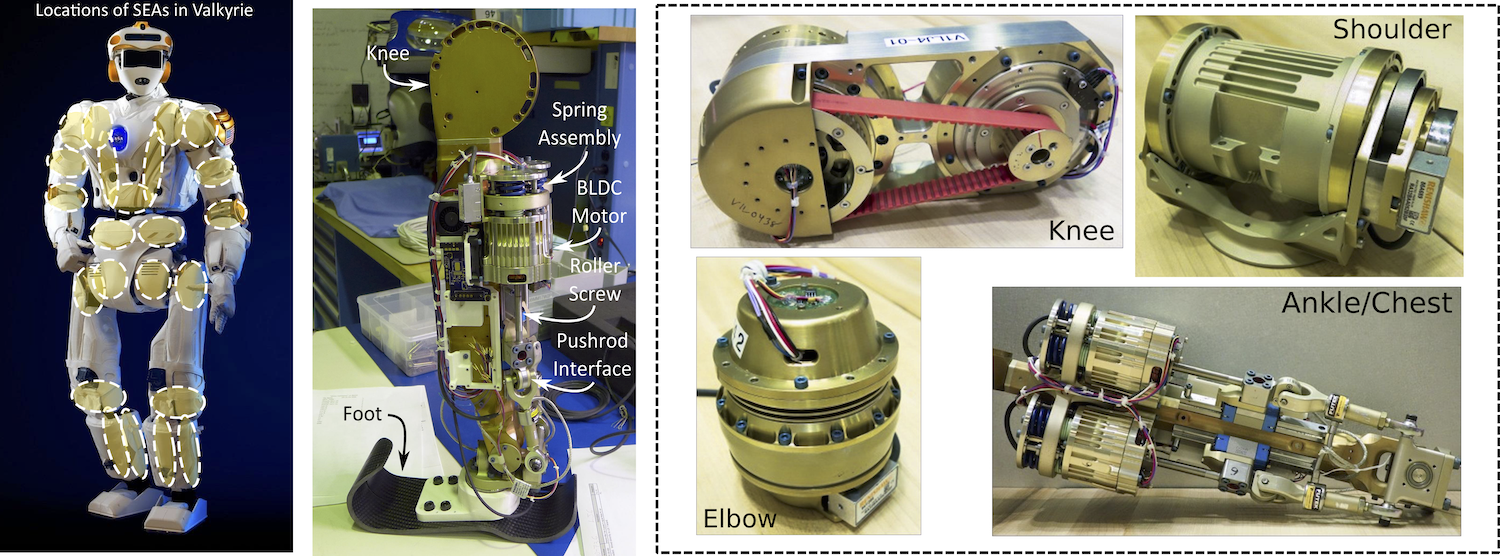}
 \caption{\captionsize Valkyrie robot equipped with series elastic actuators. The top figure shows a set of high-performance NASA Valkyrie series elastic actuators (SEAs), the bottom left one shows the Valkyrie robot with SEA location annotations and the bottom right one shows the calf and ankle structure.}
 \label{fig:SEA}
\end{figure}

Series elastic actuators \cite{pratt2004late, painedesign2014, vallery2008compliant, lu2015design, luo2018variable}, as an emerging actuation mechanism, provide considerable advantages in compliant and safe environmental interactions, impact absorption, energy storage and force sensing. In the control literature, adopting cascaded impedance control architectures for series elastic actuators (SEAs) has attracted increasing investigations over the last few years \cite{mosadeghzad2012comparison, vallery2008compliant, tagliamonterendering}. Compared to full-state feedback control \cite{albu2007unified, hutter2013efficient, de2005pd}, the cascaded control performs superior when the controlled plant comprises slow dynamics and fast dynamics simultaneously. In this case, the inner fast control loop isolates the outer slow control loop from nonlinear dynamics inherent to the physical system, such as friction and stiction. Therefore, this study focuses on the cascaded control structure to simulate the distributed control structure for humanoid robots accompanied with a variety of delayed feedback loops \cite{fok2016controlit, sakagami2002intelligent}. This class of cascaded control structures nests feedback control loops \cite{mosadeghzad2012comparison, vallery2008compliant}, i.e., an inner-torque loop and an outer-impedance loop for the task-level control, such as Cartesian impedance control. Recently, the works in \cite{vallery2008compliant, tagliamonterendering} proposed to embed a motor velocity loop inside the torque feedback loop. This velocity feedback enables to use integral gains for counteracting static errors such as drivetrain friction, while maintaining the system's passivity. The authors in \cite{mosadeghzad2012comparison} extensively studied the stability, passivity and performance for a variety of cascaded feedback control schemes incorporating position, velocity and torque feedback loops.

Robustness and effects of delay have often been studied in work regarding Proportional-Integral-Derivative (PID) controller tuning. A survey of PID controllers including system plants using phase margin techniques with linear approximations is conducted in \cite{lee-pid}. The works \cite{astrom-pid, poulin-pid} study auto-tuning and adaption of PID controllers while the work \cite{yaniv-good} furthers these techniques by developing optimal design tools applied to various types of plants which include delays. The study in \cite{tipsuwan2004gain} proposed an optimal gain scheduling method for DC motor speed control with a PI controller. In \cite{LeeJC14}, a backstepping controller with time-delay estimation and nonlinear damping is considered for variable PID gain tuning under disturbances. The high volume of studies on PID tuning methods highlight the importance of this topic for robust control under disturbances. However, none of those studies considers the sensitivity discrepancy to latencies between the stiffness and damping servos as separate entities nor do they consider the decoupling of those servos into separate processes for stability purposes as it is done in this chapter.

Optimal controller design methodologies are increasingly sought within the robotics and control community. Recent works in \cite{yeimpedance14} devised a critically-damped controller gain design criterion to accomplish high impedance for rigid actuators. However, inherent fourth-order SEA dynamics in this study make it challenging to design optimal controllers of the cascaded feedback structure.
For the cascaded control, a common routine is to tune the inner-loop gains first, followed by an outer-loop gain tuning. Indeed, this procedure consumes substantially hand-tuning efforts and lacks optimal performance guarantees. The majority of existing results rely on empirical tuning \cite{pratt2004late, mosadeghzad2012comparison}. The work in \cite{vallery2008compliant} designed controller gain ranges according to a passivity criterion. However, gain parameters were highly coupled as a set of inequalities, which leaves the controller gains undetermined. 
In this chapter, a fourth-order gain design criterion is proposed by simultaneously solving SEA optimal impedance gains and torque gains. The ``optimality" is proposed according to phase-margin-based stability. Through this criterion, the designer only needs to specify a natural frequency parameter, and then all the impedance and torque gains are deterministically solved. A larger natural frequency represents larger impedance and torque controller gains. This dimensionality reduction and automatic solving process is not only convenient for SEA controller design but also warrants optimal performance in terms of system closed-loop stability.
%

System passivity criteria have been extensively studied for coupled systems \cite{hulin2006stability, focchi2016robot, albu2007unified}, networked control systems \cite{gao2007passivity} and coordination control \cite{yin2017coordination}. Among the robotics community, the authors in \cite{vallery2008compliant} designed passivity-based controller gains for series elastic actuators. However, that work only incorporates stiffness feedback, and the ignored damping feedback indeed plays a pivotal role, which will be analyzed in this study. Damping-type impedance control was investigated in \cite{tagliamonterendering}. However, it does not analyze the effects of time delays and filtering. Although these practical issues were tackled in \cite{vallery2008compliant}, the time delays are so subtle that it can not model large time delays often existing in serial communication channels. Due to the destabilizing effects of time delays, significant effort has been put forth to ensure that systems are stable, by enforcing passivity criteria \cite{colgate1994passivity}.
In light of these discussions, the contributions of this chapter are: (i) analyze, provide control system solutions, implement and evaluate actuators and mobile robotic systems with latency-prone distributed architectures to significantly enhance their stability and trajectory tracking capabilities; (ii) analyzing time domain controller stability of series elastic actuators (SEAs) and proposing a critically-damped gain selection criterion; (iii) conducting a frequency-dependent impedance analysis of SEAs affected by time delays and filtering; We expect this study provides a promising solution of designing optimal impedance controllers for SEA-equipped humanoid robots (see Fig.~\ref{fig:SEA}) to achieve complex locomotion and manipulation tasks. 
%
The results presented in this chapter have been published in \cite{yeimpedance14, zhao2014sensitivity, zhao2018impedance, zhao2014feedback}.

\section{Modeling of Series Elastic Actuators}
\label{sec:SEAModel}
\begin{figure}
 \centering
   \includegraphics[width=0.4\linewidth]{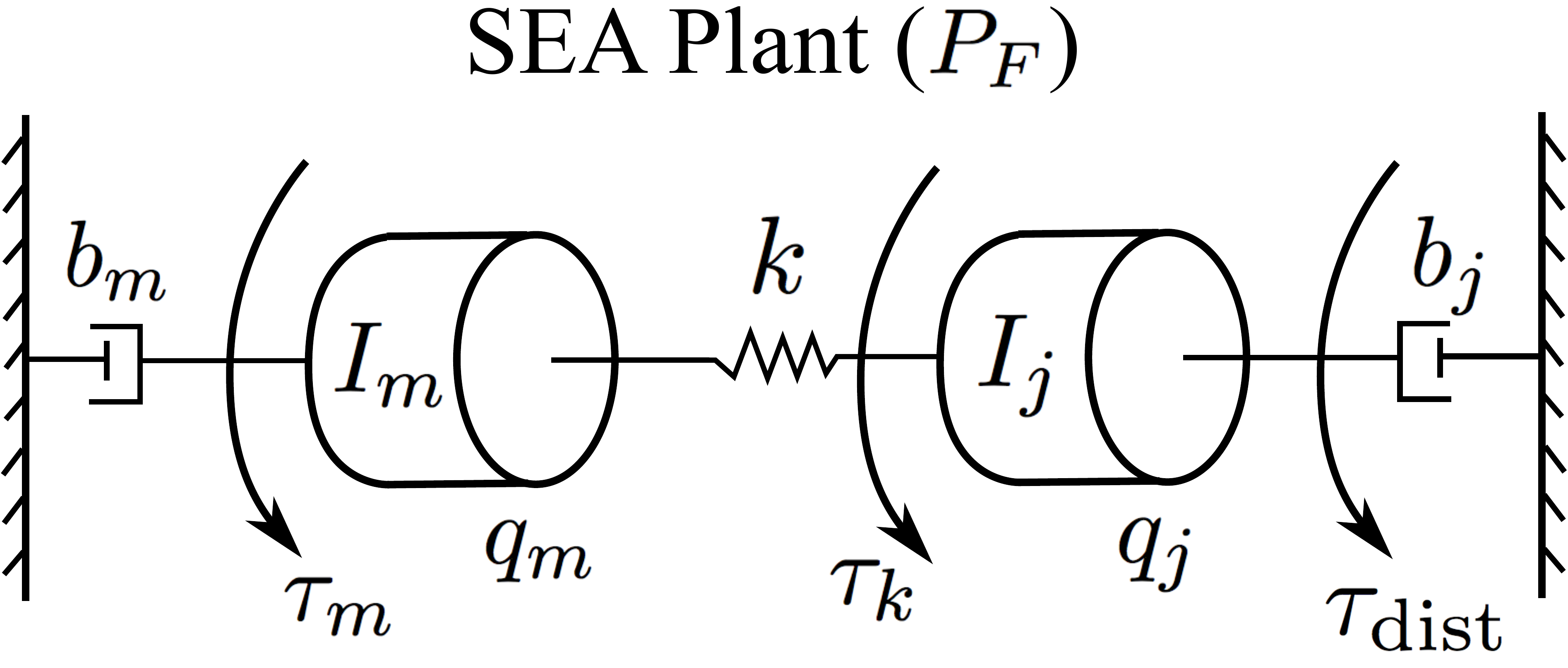}
\caption{\captionsize SEA model. The annotated parameters are defined in Section~\ref{sec:SEAModel}. We map the motor inertia $I_m$ and motor damping $b_m$ to the joint coordinates by multiplying by the gear reduction squared.}
\vspace{-0.1in}
 \label{fig:SEABlock}
\end{figure}

\begin{figure}
 \centering
   \includegraphics[width=0.7\linewidth]{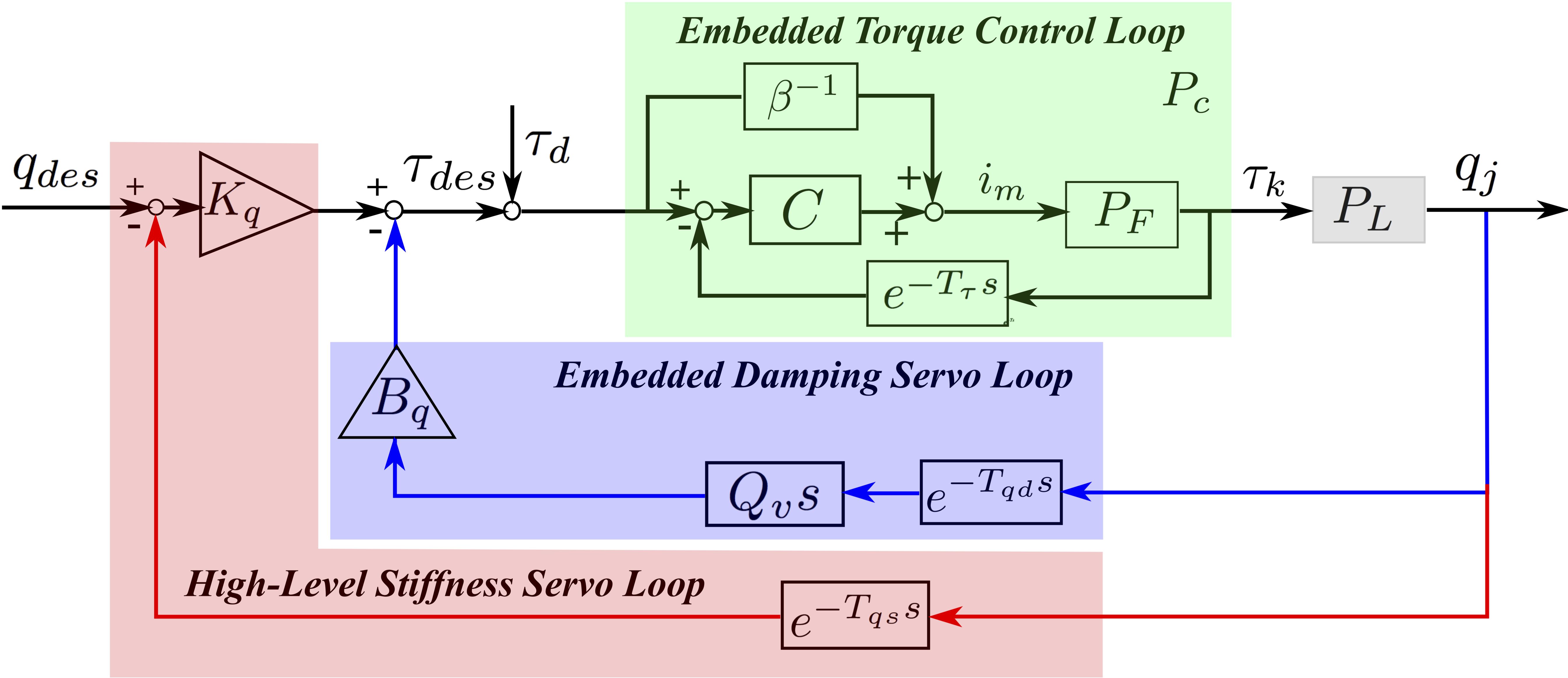}
\caption{\captionsize SEA controller diagram. The inner-torque controller is composed of a feedforward loop with a mapping scalar $\beta^{-1}$ and PD torque feedback loops. The outer-impedance controller constitutes stiffness and damping feedback loops. Time delays are modeled as $e^{-Ts}$. We apply first-order low-pass filters to both velocity and torque derivative feedback loops. 
$\tau_k$ represents the spring torque. The motor current input is $i_m$.  The embedded torque control loop is denoted by $P_C$, which normally has faster dynamics than the outer one.
}
 \label{fig:ControlBlock}
\end{figure}

This section models a series elastic actuator (SEA) constituting two nested feedback loops, i.e., an outer-impedance loop and an inner-torque loop. The SEA dynamics can be modeled as shown in Fig.~\ref{fig:SEABlock}. The spring torque $\tau_k$ is 
\begin{equation}\label{eq:springforce}
\tau_k = k (q_m - q_j).
\end{equation}
where the spring stiffness is denoted by $k$. $q_m$ and $q_j$ represent motor and joint positions, respectively. As to the joint side, it is assumed that disturbance torque $\tau_{\rm dist} = 0$. Namely, spring torque is equal to load torque, i.e., 
\begin{equation}\label{eq:loadforce}
\tau_k = I_j \ddot{q}_j + b_j \dot{q}_j.
\end{equation}
where $I_j$ and $b_j$ are joint inertia and damping coefficients, respectively. Notably, this model merely models the effects of viscous friction; we leave the analysis of other types of friction for future work. Then the load plant $P_L(s)$ has
\begin{equation}\label{eq:PL}
P_L(s) = \frac{q_j (s)}{\tau_k (s)} = \frac{1}{I_j s^2 + b_j s}.
\end{equation}
By Eqs. (\ref{eq:springforce}) and (\ref{eq:loadforce}), the following transfer function can be derived
\begin{equation}\label{eq:position}
\frac{q_j(s)}{q_m(s)} = \frac{k}{I_j s^2 + b_j s + k}.
\end{equation}
We have motor torque $\tau_m = I_m \ddot{q}_m + b_m \dot{q}_m + k (q_m - q_j)$. Combining the equation above with (\ref{eq:position}) and defining the spring deflection as $\Delta q = q_m - q_j$, we establish the following mapping from the motor angle $q_m$ to $\Delta q$
\begin{equation}
r(s) = \frac{\Delta q(s)}{q_m (s)} = \frac{I_j s^2 + b_j s}{I_j s^2 + b_j s + k}.
\end{equation}
By (\ref{eq:springforce}), we can express the spring torque as
\begin{equation}\label{eq:springforce2}
\tau_k(s) = k \Delta q(s) = k r(s) q_m(s).
\end{equation}
Given the relationship between the motor current $i_m$ and the motor torque $\tau_m$ represented by $\tau_m(s)/i_m(s) = \beta = \eta N k_{\tau}$, with drivetrain efficiency $\eta$ (constant for simplicity, and dynamic modeling of drivetrain losses is ignored), gear speed reduction $N$ and motor torque constant $k_{\tau}$ ($N$ represents the ratio of motor rotary velocity to actuator linear velocity. This gear ratio is achieved by using pulley reduction $N_p$ and ball screw, which is parameterized by ball screw lead $l_{bs}$. Please refer to \cite{painedesign2014} for more actuator design details.) See Table~\ref{table:SEAParams} in Section~\ref{sec:experiment} for more parameter details, the SEA plant $P_F(s)$ is represented by 
\begin{equation}\label{eq:PF}
P_F(s) = \frac{\tau_k (s)}{i_m (s)} = \frac{\beta r(s) k}{I_m s^2 + b_m s + r(s) k}.
\end{equation}
where $I_m$ and $b_m$ are motor inertia and damping coefficients, respectively. By Fig. \ref{fig:ControlBlock}, the closed-loop torque control plant $P_{C}$ is
\begin{equation}\label{eq:P_CF}
P_{C}(s) = \frac{\tau_k (s)}{\tau_{\rm des} (s)} = \frac{P_F(\beta^{-1}+C)}{1 + P_F C e^{-T_\tau s}}.
\end{equation}

The torque feedback loop includes a delay term $e^{-T_{\tau} s}$ and a PD compensator $C = K_{\tau} + B_{\tau} Q_{\tau d} s$ (see Fig.~\ref{fig:ControlBlock}), where $Q_{\tau d}$ models a first-order low-pass filter for the torque derivative signal,
\begin{equation}\label{eq:LPfiltertorque}
Q_{\tau d} = \frac{2 \pi f_{\tau d}}{s + 2 \pi f_{\tau d}},
\end{equation}
where $f_{\tau d}$ is the filter cut-off frequency. Additionally, a feedforward loop is incorporated to convert the desired torque $\tau_{\rm des}$ to the motor current $i_m$ (see Fig. \ref{fig:ControlBlock}). By Eqs. (\ref{eq:PL}) and (\ref{eq:P_CF}), the following transfer function can be obtained
\begin{equation}\label{eq:forceratio}
\frac{q_j(s)}{\tau_{\rm des}(s)} = P_L P_{C} = \frac{P_F (\beta^{-1} + C)}{(1 + P_F C e^{-T_\tau s}) (I_js^2 + b_j s)}.
\end{equation}
\begin{figure}[t]
 \centering
   \includegraphics[width=0.95\linewidth]{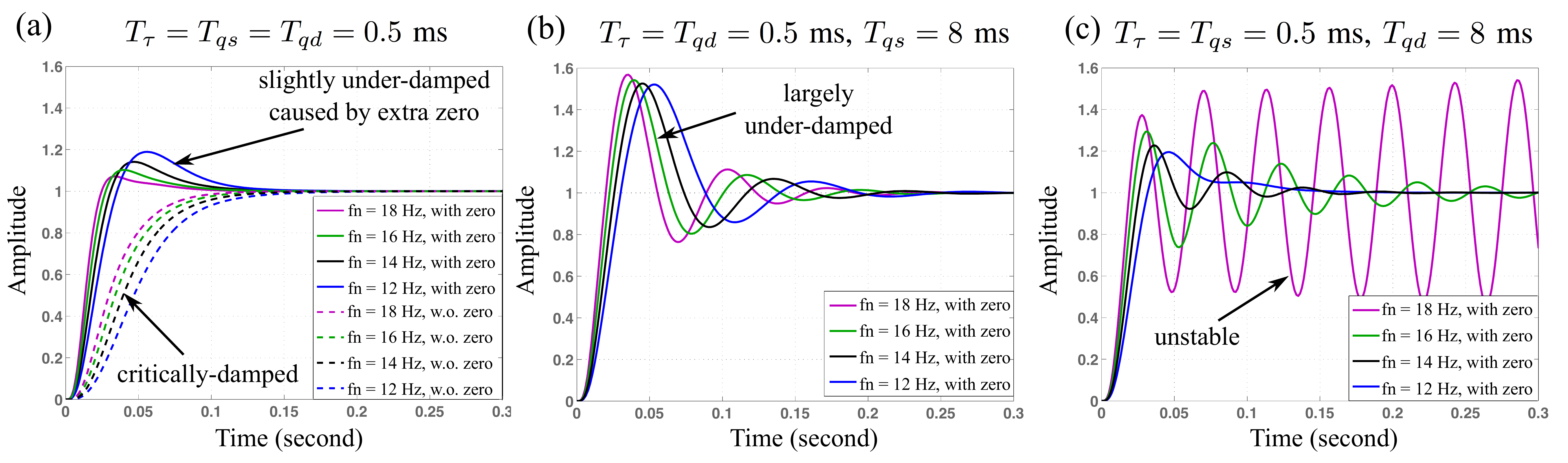}
 \caption{\captionsize SEA step response affected by time delays. These subfigures demonstrate that the larger delays that impedance feedback loops have, the worse performance that the step response has. As shown in subfigures (b) and (c), SEA stability has a higher sensitivity to damping delays than the stiffness counterpart. Subfigure (a) reveals that larger $f_n$ leads to a larger overshoot, which appears to be counterintuitive. However, by close inspection, we can observe that the largest $f_n$ in the solid magenta color already shows distortion, and its $36.4^\circ$ phase margin is the smallest among all four cases. To study the influence of zero in (\ref{eq:closedloop}), step responses without this zero are also simulated and represented by dashed lines in (a). By comparison, we can realize an overshoot induced by this extra zero.}
\label{fig:ImpedanceDelay}
 \vspace{-0.15in}
\end{figure}
For the impedance feedback, we have the form as below
\begin{equation}
\tau_{\rm des}(s) = K_{q} (q_{\rm des} - e^{-T_{qs} s} q_j) - B_{q} e^{-T_{qd} s} Q_{qd} s q_j,
\end{equation}
where $e^{-T_{qs}s}$ and $e^{-T_{qd}s}$ denote the time delays of stiffness and damping feedback loops, separately. The joint velocity filter $Q_{qd}$ has the same format as that in (\ref{eq:LPfiltertorque}) with a cut-off frequency $f_{qd}$. Alternatively, we can also send the desired joint velocity as the input of the embedded damping loop. In that case, an extra zero will show up in the numerator of (\ref{eq:closedloop}). Since a zero only changes transient dynamics, it does not affect system stability. Using $P_L$ and $P_{C}$ in Eqs. (\ref{eq:PL}) and (\ref{eq:P_CF}), we obtain the SEA closed-loop transfer function $P_{CL}$,
\begin{align}\label{eq:closedloop}
P_{CL}(s) = \frac{q_j (s)}{q_{\rm des} (s)} = \frac{K_{q}P_{C} P_L }{1 + P_{C} P_L (e^{-T_{qd} s} B_{q} Q_{qd} s + e^{-T_{qs} s} K_{q})} = \frac{K_{q}(1 + \beta K_{\tau} + \beta B_{\tau} Q_{\tau d}s)}{\sum\nolimits_{i = 0}^4 D_{i} s^i},
\end{align}
with the associated coefficients defined as


\begin{align}\nonumber
D_4 = & I_m I_j/k, D_3 = (I_j b_m + I_m b_j)/k + I_j \beta B_{\tau} Q_{\tau d} e^{-T_{\tau} s},\\\nonumber
D_2 = & I_j (1 + e^{-T_{\tau}s} \beta K_{\tau}) + I_m + b_j \beta B_{\tau} Q_{\tau d}e^{-T_{\tau} s} + \beta B_{\tau} B_{q} e^{-T_{qd} s} Q_{qd}Q_{\tau d} + b_j b_m/k, \\\nonumber
D_1 = & b_j (1 + e^{-T_{\tau}s} \beta K_{\tau}) + b_m + \beta B_{\tau} Q_{\tau d}K_{q} e^{-T_{qs} s} + e^{-T_{qd} s} (1 + \beta K_{\tau}) B_{q} Q_{qd},  \\
D_0 = & e^{-T_{qs} s} (1 + \beta K_{\tau}) K_{q}.
\end{align}

This closed-loop transfer function is sixth-order due to the existence of low-pass filters $Q_{qd}$ and $Q_{\tau d}$. Here we formulate it in fourth-order form for the sake of clarity. Note that, the numerator of (\ref{eq:closedloop}) has a zero, induced by the torque derivative term. As to the step response, this induced zero shortens the rise time but causes an overshoot. Nevertheless, system stability is not affected since it is solely determined by the denominator's characteristic polynomial. 
\section{Gain Design of series elastic actuators}
\label{sec:gainselection}
The closed-loop transfer function derived in (\ref{eq:closedloop}) is complex due to the cascaded impedance and torque feedback loops. This complexity makes the SEA controller design challenging. In this section, we propose a critically-damped criterion to design optimal controller gains.
\subsection{Critically-damped controller gain design criterion} 
Impedance control gains of rigid actuators can be designed based on the well-established critically-damped criterion of second-order systems \cite{yeimpedance14}. As for high-order systems like SEAs, such a critically-damped criterion is still missing. In this study, we aim at designing feedback controller gains such that the overall SEA closed-loop system behaves as two damped second-order systems \cite{petit2011state}. To this end, we represent the fourth-order system in (\ref{eq:closedloop}) (the time delays and filtering in (\ref{eq:closedloop}) are ignored for problem tractability) by two second-order systems in multiplication presented as 
\begin{equation}\label{eq:standardfourthorder}
(s^2 + 2\zeta_1 \omega_1 s + \omega_1^2)(s^2 + 2\zeta_2 \omega_2 s + \omega_2^2),
\end{equation}
which has four design parameters $\omega_1, \omega_2, \zeta_1, \zeta_2$. They will be used to design the gains $K_{q}, B_{q}, K_{\tau}$, and $B_{\tau}$. First, we set $\zeta_1 = \zeta_2 = 1$ in (\ref{eq:standardfourthorder}) to obtain the critically-damped performance. Second, we assume $\omega_2 = \omega_1$ for simplicity. An optimal pole placement design is left for future work. Let us define a natural frequency $f_n$ of (\ref{eq:standardfourthorder}) as 
\begin{equation}\label{eq:naturalFreq}
\omega_1 = \omega_2 \triangleq \omega_n = 2 \pi f_n.
\end{equation}
By comparing the denominators of Eqs. (\ref{eq:closedloop}) and~(\ref{eq:standardfourthorder}), we obtain the nonlinear gain design criterion equations as shown below

\begin{align}\nonumber
&\frac{I_j b_m + I_m b_j + I_j \beta B_{\tau} k}{I_m I_j} = 4 \omega_n, \\\nonumber
&\frac{k (I_j (1 + \beta K_{\tau}) + I_m + \beta B_{\tau}(b_j + B_{q})) + b_j b_m}{I_m I_j} = 6 \omega_n^2, \\\nonumber
&\frac{k (b_j + B_{q})(1 + \beta K_{\tau}) + k (b_m + \beta B_{\tau}K_{q})} {I_m I_j} = 4 \omega_n^3, \\\label{eq:nonlinearEqs}
&\frac{(1 + \beta K_{\tau}) k K_{q}}{I_m I_j} = \omega_n^4.
\end{align}
These four equations with coupled gains can be solved by Matlab's fsolve() function. Note that, representing a fourth-order system by two multiplied second-order systems in (\ref{eq:standardfourthorder}) maintains the properties of the fourth-order system. In our method above, the simplification comes from the selection of $\omega_1, \omega_2, \zeta_1, \zeta_2$ parameters in (\ref{eq:standardfourthorder}). The resulting benefit is that selecting a natural frequency uniformly determines all the gains of torque and impedance controllers. This advantage avoids the commonly-adopted complicated yet heuristic controller tuning procedures, like the ones in \cite{mosadeghzad2012comparison, petit2011state}, although system dynamics in our case are restricted to specific patterns such as the critically-damped one we design. Let us show an example as follows.
\begin{example}
To validate this criterion, we test five natural frequencies. We select filter cut-off frequencies $f_{vd} = 50$ Hz, $f_{\tau d} = 100$ Hz and time delays $T_\tau = T_{qd} = 0.5$ ms, $T_{qs} = 2$ ms. These filters and delays are only used in the phase-space computation based on (\ref{eq:closedloop}), and ignored in the critically-damped selection criterion for problem tractability. The solved gains and phase margins are shown in Table~\ref{table:CriticalDampedGainSet}. Noteworthily, the phase margin is computed based on the open-loop transfer function derived from $P_{CL}$ in (\ref{eq:closedloop}). Increasing $f_n$ will lead to a uniform increase of all four gains. This property meets our expectation that increasing torque (or impedance) gains results in a torque (or impedance) bandwidth increase and a phase margin decrease. 
\end{example}
\begin{table}[t]
\caption{Critically-damped Controller Gains} 
\centering
\begin{tabular}{c||c|c|c}
\hline
Frequency & Impedance Gains & Torque Gains & Phase\\ 
(Hz) & (Nm/rad, Nms/rad) & (A/Nm, As/Nm) & Margin \\ \hline\hline
\multirow{2}{*}{$f_n = 12$} & $K_q = 65$ & $K_{\tau} = 1.18$ & \multirow{2}{*}{$45.1^\circ$}\\
& $B_q = 0.46$ & $B_{\tau} = 0.057$ &\\ \hline
\multirow{2}{*}{$f_n = 14$} & $K_q = 83$ & $K_{\tau} = 1.80$ & \multirow{2}{*}{$43.2^\circ$}\\
& $B_q = 0.76$ & $B_{\tau} = 0.067$ &\\ \hline
\multirow{2}{*}{$f_n = 16$} & $K_q = 103$ & $K_{\tau} = 2.56$ & \multirow{2}{*}{$40.0^\circ$}\\
& $B_q = 1.02$ & $B_{\tau} = 0.077$ &\\ \hline
\multirow{2}{*}{$f_n = 18$} & $K_q = 124$ & $K_{\tau} = 3.45$ & \multirow{2}{*}{$36.5^\circ$}\\
& $B_q = 1.26$ & $B_{\tau} = 0.087$ &\\ \hline
\multirow{2}{*}{$f_n = 20$} & $K_q = 148$ & $K_{\tau} = 4.48$ & \multirow{2}{*}{$33.2^\circ$}\\
& $B_q = 1.49$ & $B_{\tau} = 0.097$ &\\ \hline
\end{tabular}
\label{table:CriticalDampedGainSet}
\vspace{-3mm}
\end{table}

Note that, for simplicity, the gain design above ignores time delay, which does affect system stability. Next, we will study the effect of time delays given this gain design criterion. Since torque feedback is the inner loop, it normally suffers a smaller delay than that in the outer impedance loop. This is why we assign $T_{\tau} = 0.5$ ms in the example above. Notice that $T_{qs}$ is chosen to be larger than $T_{qd}$ since the former belongs to the outer control loop while the latter belongs to the inner control loop. The benefits of having damping feedback in the inner loop was extensively analyzed in \cite{yeimpedance14}.
This motivates us to implement the impedance feedback loops in a distributed pattern as shown in Fig.~\ref{fig:ControlBlock}. Namely, we allocate the stiffness feedback loop at the high level while embedding the damping feedback loop at the low level for a fast servo rate. The same distributed control strategy was implemented for the rigid actuators in \cite{yeimpedance14} and extended lately for the Whole-Body Operational Space Control \cite{zhao2016passivity, zhao2017planning}.

\subsection{Trade-off between torque and impedance control}
During gain tuning of the SEA-equipped bipedal robot Hume and NASA Valkyrie robot, which have similar SEA control architectures as the one in Fig.~\ref{fig:ControlBlock}, a pivotal phenomenon is observed: if one increases torque controller gains or decreases impedance controller gains, the robot tends to become unstable. To reason about this observation, we propose a SEA gain scale definition as follows
\begin{definition}[SEA Gain Scale]\label{def:gainscale}
The gain scale of a SEA's cascaded controller is a scaling parameter $GS$ between adjusted gains ($K_{i_a}, B_{i_a}$) and nominal gains ($K_{i_n}, B_{i_n}$), $i \in \{\tau, q\}$,
\begin{align}\label{eq:gainscale}
GS = \frac{K_{\tau_a}}{K_{\tau_n}} = \frac{K_{q_n}}{K_{q_a}}, \quad GS = \frac{B_{\tau_a}}{B_{\tau_n}} = \frac{B_{q_n}}{B_{q_a}},
\end{align}
where the adjusted gains denote actual gains in use while the nominal gains denote reference ones designed by the critically-damped gain design criterion.
\end{definition}
It should be noted that if $GS = 1$, then the adjusted gains are the same as the nominal gains. For example, the controller gains in Table~\ref{table:CriticalDampedGainSet} are five sets of nominal gains.
\begin{figure}
 \centering
   \includegraphics[width=0.7\linewidth]{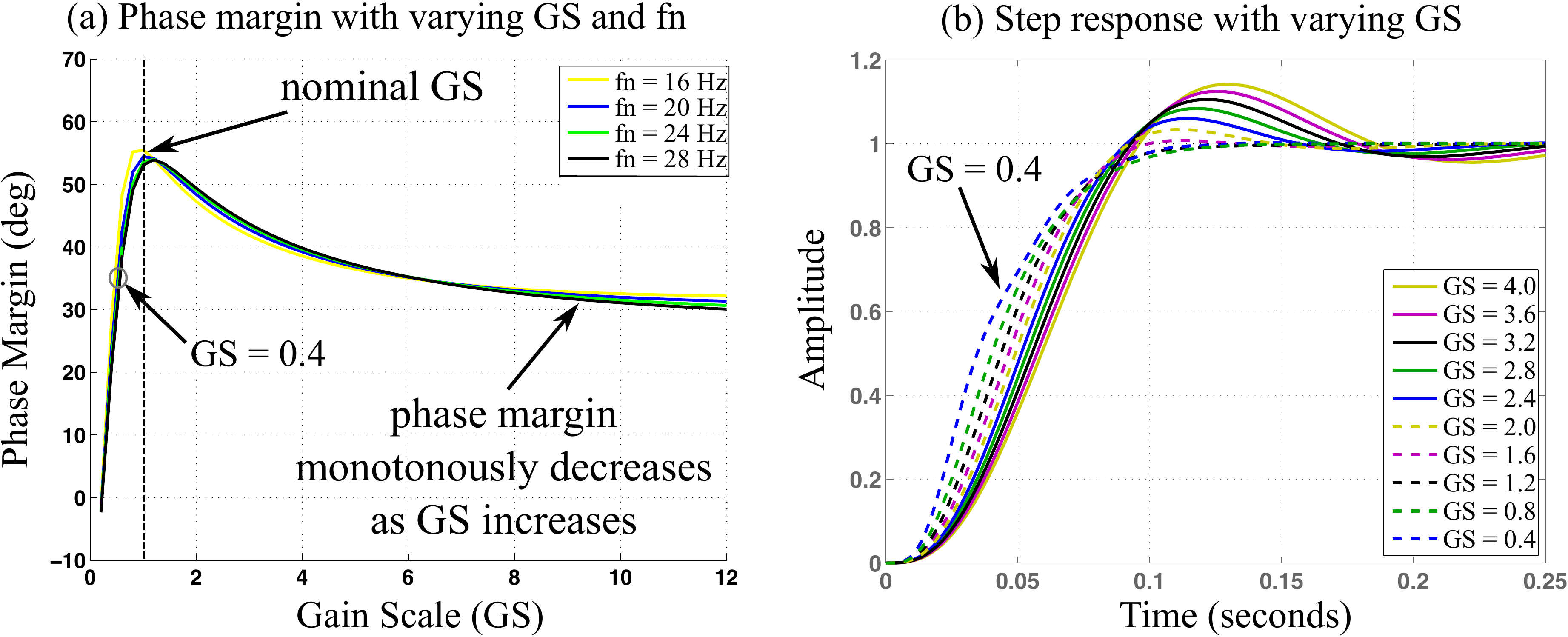}
 \caption{\captionsize Optimality of the critically-damped gain design criterion. Subfigure (a) samples a variety of gain scales and natural frequencies. An optimal performance is achieved by using the proposed critically-damped gain design criterion. Subfigure (b) shows: (i) a larger overshoot but slow rise time when $GS > 1$; (ii) an over-damped response with distortions when $GS < 1$.}
 \label{fig:GainScale}
\vspace{-0.05in}
\end{figure}
By (\ref{eq:gainscale}), we have the following equalities
\begin{align}\label{eq:multipliedgains}
K_{\tau_a} \cdot K_{q_a} = K_{\tau_n} \cdot K_{q_n}, \quad B_{\tau_a} \cdot B_{q_a} = B_{\tau_n} \cdot B_{q_n},
\end{align}
which maintains the same multiplicative value of nested proportional (or derivative) torque and impedance gains for the normal and adjusted conditions. An overall controller gain design procedure is shown in Algorithm~\ref{al:controllerDesign}.
\begin{algorithm}[t]
\caption{Gain controller design procedure}\label{al:controllerDesign}
\begin{algorithmic}
\State Assign system parameters \textit{sysParam} in (\ref{eq:closedloop}).
\State Assign natural frequency $f_n$ (i.e., $\omega_1$ and $\omega_2$ by (\ref{eq:naturalFreq})), $\zeta_1 = \zeta_2 \gets 1$. 
\Procedure{ControllerSolver}{$f_n, \zeta_1, \zeta_2$, \textit{sysParam}}
\State Deterministically solve nominal controller gains $K_{q_n},$ 
\State $B_{q_n}, K_{\tau_n}, B_{\tau_n}$ \Comment{refer to (\ref{eq:nonlinearEqs})} 
\If {Gain scale $GS = 1$}
\State $(K_{q}, B_{q}, K_{\tau}, B_{\tau}) \gets (K_{q_n}, B_{q_n}, K_{\tau_n}, B_{\tau_n})$
\Else {}
\State $(K_{q_a}, B_{q_a}) \gets (K_{q_n}, B_{q_n})/GS$
\State $(K_{\tau_a}, B_{\tau_a}) \gets GS \cdot(K_{\tau_n}, B_{\tau_n})$ \Comment{refer to (\ref{eq:gainscale})}
\State $(K_{q}, B_{q}, K_{\tau}, B_{\tau}) \gets (K_{q_a}, B_{q_a}, K_{\tau_a}, B_{\tau_a})$
\EndIf
\State \textbf{return} $(K_q, B_q, K_{\tau}, B_{\tau})$
\EndProcedure
\State Assign filtering parameters $f_{vd}, f_{\tau d}$ and time delays $T_{\tau}, T_{qs}, T_{qd}$.
\State PM = PhaseMargin$(K_q, B_q, K_{\tau}, B_{\tau}, f_{vd}, f_{\tau d}, T_{\tau}, T_{qs}, T_{qd})$
\end{algorithmic}
\label{algorithm:PSExecution}
\end{algorithm}

There is a trade-off between a large torque bandwidth for accurate torque tracking and a low torque bandwidth for larger achievable impedance range. The work in \cite{focchi2016robot} obtained a similar observation that enlarging the inner loop controller bandwidth reduces the range of stable impedance control gains.
In their experimental validations, they do not decrease impedance gains when raising torque gains. As it is known, the product of cascaded gains grows if torque gains increase, however this increase is not considered in their stability analysis. It is therefore unclear if the reduced stable impedance range is caused by enlarging the torque gains or the increased product gain due to the coupled effect of torque and impedance gains. To validate the trade-off in a more realistic manner, our method maintains a constant gain product value as shown in ~(\ref{eq:multipliedgains}). Fig.~\ref{fig:GainScale}(a) shows the sampling results for different gain scales GS. A larger $GS$ indicates increased torque gains with decreased impedance gains. When $GS>1$, an increasing $GS$ deteriorates the system stability (i.e., phase margin) and causes a larger oscillatory step response as shown in Fig.~\ref{fig:GainScale}(b). On the other hand, when $GS < 1$, a decreasing $G$ also decreases the system stability. For instance, $GS = 0.4$ corresponds to a $34^\circ$ phase margin as shown in subfigure (a), and accordingly a distortion appears in the step response of subfigure (b). We ignore delays and filtering to focus on the effects of the gain scale. The tests in Fig.~\ref{fig:GainScale} validate the optimal performance (i.e., maximized phase-margin) of our proposed critically-damped gain design criterion (i.e., $GS = 1$). Although $GS = 1$ is the optimal value for stability, changing $GS$ to different values allows to change the impedance behavior without changing the natural frequency. Thus, we assign $GS$ as a design parameter in Algorithm~\ref{al:controllerDesign}. In the next section, we will analyze the frequency-domain SEA impedance.
\begin{figure*}
 \centering
   \includegraphics[width=0.9\linewidth]{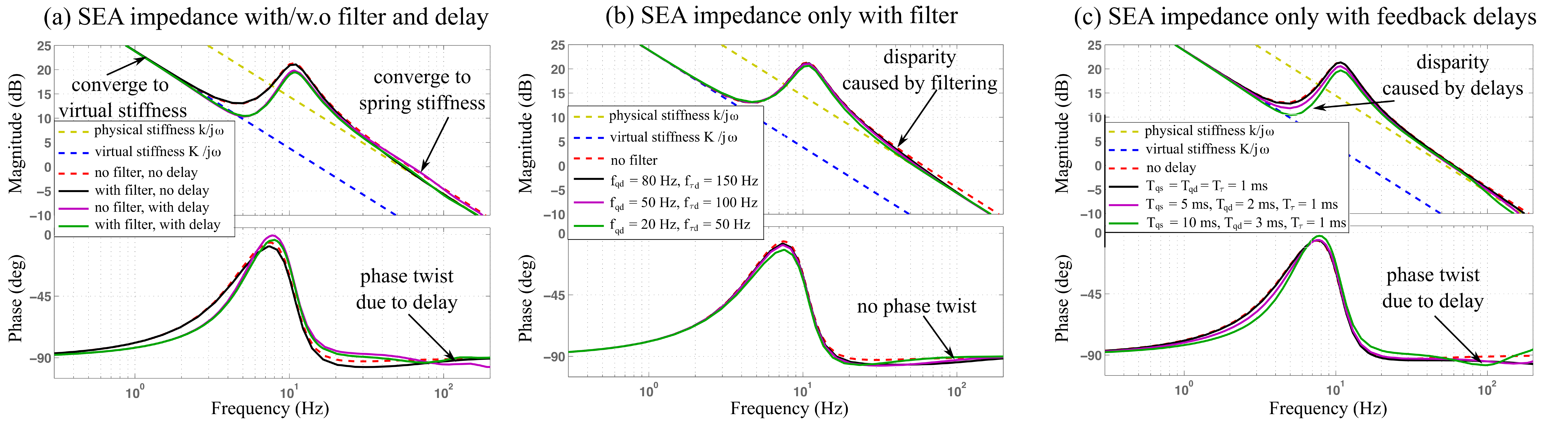}
 \caption{\captionsize SEA impedance with time delays and filtering. In subfigure (a),  the impedance of a physical spring $k/(j\omega)$ and a virtual stiffness gain controller are shown by yellow and blue dashed lines, respectively. The ideal SEA impedance without delay and filtering is represented by a red dashed line. At low frequency range, SEA impedance converges to the virtual stiffness. A similar behavior was observed in \cite{vallery2008compliant}. At high frequency range, it approaches another impedance asymptote. Subfigure (b) analyzes the filtering effect while subfigure (c) analyzes the time delay effect. Accordingly, the sensitivity discrepancy of different time delays can be analyzed but not discussed here due to the space limit. These simulations have a natural frequency $f_n = 30$ Hz, corresponding to $K_q = 293.6$ Nm/rad, $B_q = 2.49$ Nms/rad, $K_\tau = 11.71$ A/Nm, $B_\tau = 0.146$ As/Nm.}
 \label{fig:impedanceFilterDelay}
\vspace{-0.1in}
\end{figure*}

\section{SEA impedance analysis}
\label{sec:SEAimpedance}
Impedance control is widely used for dynamic interaction between a robot and its physically interacting environment \cite{hogan1985impedance}. In this section, we study SEA impedance performance in the frequency domain. In particular, we first derive the SEA impedance transfer function given the SEA controller diagram in Fig.~\ref{fig:ControlBlock}, and then analyze the effects of time delays, filtering and load inertia.
\subsection{SEA impedance transfer function}
The SEA impedance transfer function is defined with a joint velocity $\dot{q}_j$ input and a joint torque $\tau_j$ output. Based on zero desired joint position $q_{\rm des}$, the SEA impedance $Z(s) = \tau_j(s)/(-sq_j(s))$ is formulated as follows
\begin{equation}\label{eq:SEA_impedance}
Z(s) = \frac{\tau_j(s)}{-sq_j(s)} = \frac{\sum\nolimits_{i = 0}^4 N_{zi} s^i}{
\sum\nolimits_{i = 0}^5 D_{zi} s^i},
\end{equation}
with the numerator coefficients,
%
\begin{align}\nonumber
N_{z4} = \;& I_m  T_{f\tau} T_{fv} \beta k, \\\nonumber
N_{z3} = \;& \beta k (I_m  (T_{f\tau} + T_{fv}) + T_{f\tau} T_{fv}b_m),
\\\nonumber
N_{z2} = \;& I_m  \beta k  + \beta k b_m(T_{f\tau}  + T_{fv}) + k k_\tau  (T_{f\tau} +  \beta(B_\tau + K_\tau T_{f\tau})) (B_q e^{-T_{qd}s} + K_q T_{fv} e^{-T_{qs}s}),\\\nonumber 
N_{z1} = \;& b_m \beta k  + B_q k k_\tau (1 + K_\tau \beta) e^{-T_{qd}s} + K_q k k_\tau (T_{fv} + T_{f\tau} + \beta (B_\tau + K_\tau (T_{f\tau} + T_{fv})))  e^{-T_{qs}s}, \\\nonumber
N_{z0} = \;& K_q k k_\tau e^{-T_{qs}s} (1 + K_\tau \beta),
\end{align}
and the denominator coefficients,
%
%
\begin{align}\nonumber
D_{z5} = \;& I_m T_{f\tau} T_{fv} \beta, D_{z4} = I_m \beta (T_{fv}  + T_{f\tau}) + T_{fv} T_{f\tau} \beta b_m,\\\nonumber
D_{z3} = \;& \beta I_m + \beta b_m(T_{f\tau} + T_{fv}) + T_{fv} k \beta  (T_{f\tau} + k_\tau(B_\tau + K_\tau T_{f\tau})e^{-T_\tau s}), \\\nonumber
D_{z2} = \;& \beta (b_m + T_{f\tau} k + k k_\tau (B_\tau + K_\tau T_{f\tau}) e^{- T_\tau s}) + T_{fv} \beta k (1 + K_\tau k_\tau e^{- T_\tau s}), \\\nonumber
D_{z1} = \;& \beta k (1 + K_\tau k_\tau e^{- T_\tau s}), D_{z0} = 0.
\end{align}
Note that, $Z(s)$ in (\ref{eq:SEA_impedance}) does not incorporate the joint inertia $I_j$ and damping $b_j$ since these parameters belong to parts of the interacting environment. (\ref{eq:SEA_impedance}) explicitly models time delays and filtering, which are often ignored in the literature of SEA cascaded controller architectures with PD-type controllers. 
Also, the SEA transfer function in (\ref{eq:SEA_impedance}) is complete without any approximations.
\subsection{Effects of time delays and filtering}
The SEA impedance frequency responses are demonstrated in Fig.~\ref{fig:impedanceFilterDelay}. We analyze various scenarios either with or without time delays and filtering: (i) $Z_i(j\omega)$ is the ideal impedance without delays and filtering; (ii) $Z_f(j\omega)$ is the impedance only with filtering; (iii) $Z_d(j\omega)$ is the impedance only with delays; (iv) $Z_{fd}(j\omega)$ is the impedance with both delays and filtering. At low frequency range, the SEA impedance converges to a virtual stiffness asymptote in all scenarios (when time delays are considered, we have $e^{-T_{qs}j\omega}\rightarrow 1, e^{-T_{\tau}j\omega} \rightarrow 1$ as $\omega \rightarrow 0$)
\begin{equation}\nonumber
\lim_{\omega \to 0} Z_c(j\omega) = \lim_{\omega \to 0}\frac{N_{z0}}{j\omega \cdot D_{z1}} = \frac{K_q k_\tau (\beta^{-1} + K_\tau)}{j\omega \cdot (1 + K_\tau k_\tau)},
\end{equation}
where $c \in \{i, f, d, fd\}$. The denominator of the final expression has a $j\omega$ term, which indicates a $-20$ dB/dec decay rate. The low frequency impedance $Z_c(j\omega)$ behaves as a constant stiffness impedance $K_q/j\omega$ scaled by a constant $k_\tau (\beta^{-1} + K_\tau)/(1 + K_\tau k_\tau)$. This scaling applies to any PD-type cascaded impedance controller. Note that, $k_\tau \beta^{-1}$ is normally a small value. When $k_\tau K_\tau$ is large enough, $Z_c(j\omega)$ approaches $K_q/(j\omega)$, i.e., a pure virtual spring. This meets our intuition.


As to the high frequency range, the impedance also approaches an asymptote with a potential twist, depending on the delay and filtering conditions. First, let us start with the ideal case (i), i.e., without delays and filtering. This leads to $D_{z5} = D_{z4} = 0$, and we have
\begin{align}\nonumber
\lim_{\omega \to + \infty} Z_i(j\omega) = \lim_{\omega \to + \infty} \frac{N_{z2}}{j\omega \cdot D_{z3}} = \frac{k(I_m + k_\tau B_\tau B_q)}{j\omega \cdot I_m},
\end{align}
which represents a constant stiffness-type impedance scaled from the passive spring stiffness $k/(j\omega)$. The red dashed lines in Fig.~\ref{fig:impedanceFilterDelay} illustrate this ideal SEA impedance feature. 
\begin{figure}[t]
 \centering
   \includegraphics[width=0.9\linewidth]{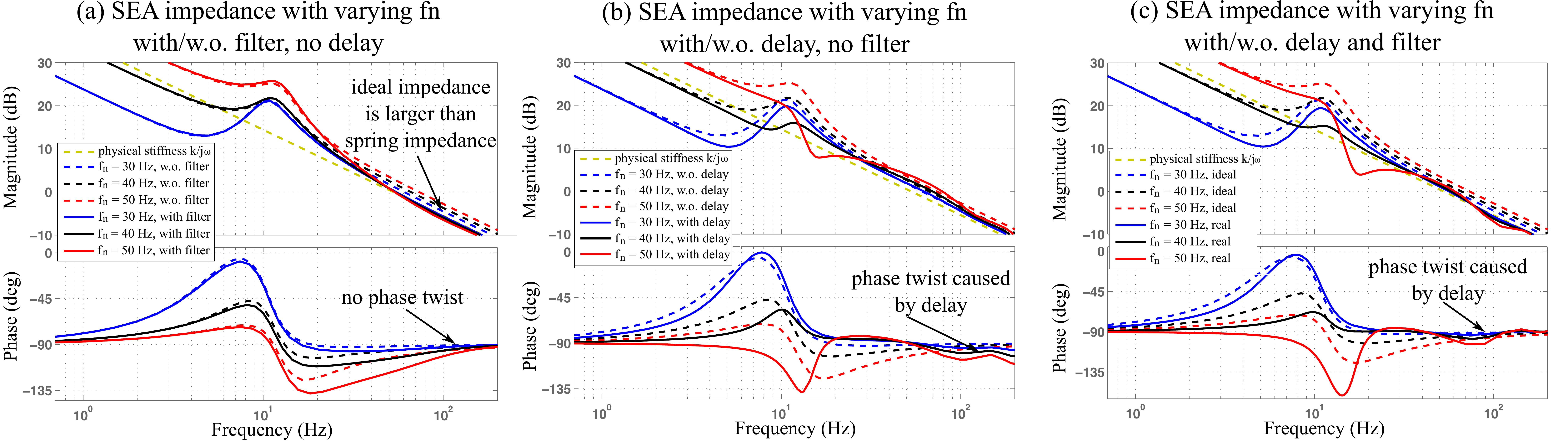}
 \caption{\captionsize SEA impedance with varying natural frequencies $f_n$. First, these subfigures validate that a higher natural frequency $f_n$ results in higher SEA impedance. Subfigures (a) and (b) show how time delay and filtering affect SEA impedance, respectively. We use filters with $f_{qd} = 50$ Hz and $f_{\tau d} = 100$ Hz while time delays are chosen as $T_{qd} = T_{\tau} = 1$ ms, $T_{qs} = 10$ ms. Second, we test the cases with both filters and delays as shown in subfigure (c), and compare them with ideal cases with neither filter nor delays.}
 \label{fig:impedanceNaturalFreq}
\end{figure}

Second, we derive the case (iii) only with delay, that is, $T_{fv} = T_{f\tau} = 0$. Then $D_{z5} = D_{z4} = 0$, and we obtain
\begin{align}\nonumber
\lim_{\omega \to + \infty} Z_d(j\omega) & = \lim_{\omega \to + \infty} \frac{N_{z2}}{j\omega \cdot D_{z3}} = \frac{k(I_m + k_\tau B_\tau B_q e^{-T_{qd}s})}{j\omega \cdot I_m}
\end{align}
Since the complex number $e^{-T_{qd}s}$ rotates along the unit circle, the SEA impedance will periodically twist around the passive spring stiffness at high frequency range. This is visualizable in Fig.~\ref{fig:impedanceFilterDelay}(c).

Third, in the case (ii) only with filtering, we have $T_{qs} = T_{qd} = T_{\tau} = 0$, and then obtain
\begin{equation}\nonumber
\lim_{\omega \to + \infty} Z_f(j\omega) = \frac{N_{z4}}{j\omega D_{z5}} = \frac{k}{j \omega},
\end{equation}
which represents a passive spring stiffness as shown in Fig.~\ref{fig:impedanceFilterDelay}(b). The curve does not twist thanks to the constant limit value $k/(j\omega)$. To verify the applicability of the behaviors aforementioned to different natural frequencies, we analyze the SEA impedance performance under varying natural frequencies in Fig.~\ref{fig:impedanceNaturalFreq}. By comparing Fig.~\ref{fig:impedanceNaturalFreq}(a) and (b) (or Fig.~\ref{fig:impedanceFilterDelay}(b) and (c)), we conclude that time delays have a larger effect on the SEA impedance than filtering.

\subsection{Effect of load inertia}
\begin{figure}[t]
 \centering
   \includegraphics[width=0.4\linewidth]{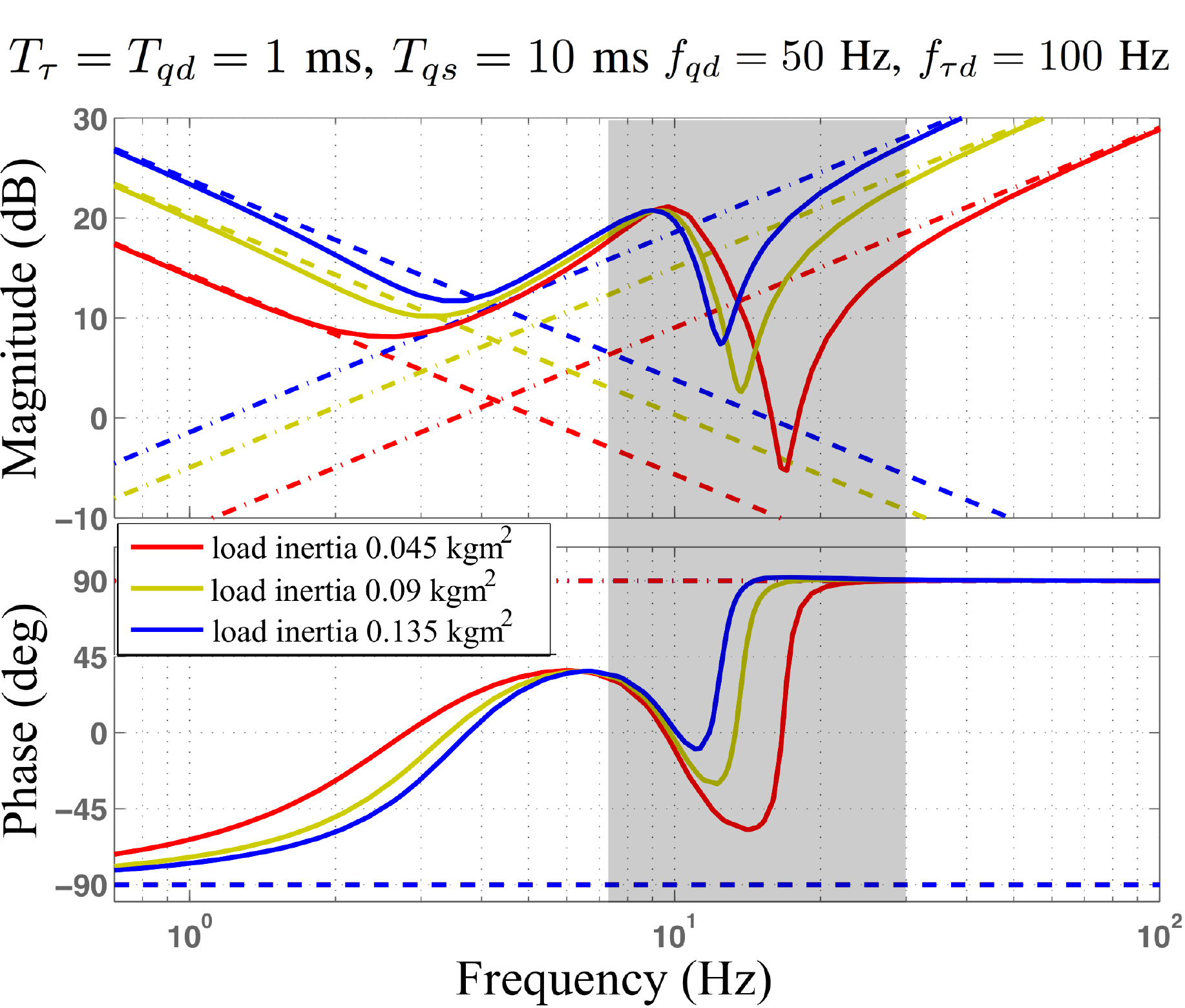}
 \caption{\captionsize SEA impedance with varying load inertias. Three different load scenarios are illustrated. All of them use the natural frequency $f_n = 20$ Hz, corresponding to $K_q = 148$ Nm/rad, $B_q = 1.49$ Nms/rad, $K_\tau = 4.48$ A/Nm, $B_\tau = 0.097$ As/Nm. The damping term is $b_j = 0.1$ Nms/rad. For all three scenarios, dashed lines are used to represent asymptote at low- and high- frequencies, respectively. Since the load inertia is modeled, the SEA impedance approaches the load inertia impedance curve $I_j \cdot j \omega + b_j$ at high frequencies.}
 \label{fig:loadinertia}
\end{figure}

This subsection analyzes the effect of load inertia on SEA impedance performance. A second-order model of the output load $I_j s + b_j$ is added into (\ref{eq:SEA_impedance}), i.e., $Z_l(j\omega) = Z(j\omega) + I_j s + b_j$. Since (\ref{eq:SEA_impedance}) becomes $Z(j\omega) \rightarrow 0$ as $\omega \rightarrow + \infty$, we have
\begin{align}\nonumber
\lim_{\omega \to + \infty} Z_l(j\omega) = \lim_{\omega \to + \infty} (Z(j\omega) + I_j \cdot j \omega + b_j) = I_j \cdot j \omega  + b_j
\end{align}
where $I_j \cdot j \omega$ represents a $20$ dB/dec asymptote at high frequencies (see Fig.~\ref{fig:loadinertia}); the damping term $b_j$ adds a constant offset. As the equation above shows, at high frequency range, SEA impedance behaves as a spring-mass impedance instead of a pure spring one. In particular, this impedance is dominated by the load inertia as shown in Fig.~\ref{fig:loadinertia}. This figure simulates three scenarios with different load inertias. Different than the load mass effect studied in \cite{vallery2008compliant}, our study has a large focus on analyzing the effect of filtering and time delays. These two factors dominate at middle frequency range where large spikes show up in the shaded region of Fig.~\ref{fig:loadinertia}. The larger load inertia is, the smaller spike the response has.

\section{Experimental Validation}
\label{sec:experiment}

\subsection{Evaluation of the controller design}
This experiment section validates the proposed methods and criterion on our series elastic actuator testbed, parameters of which are provided in Table~\ref{table:SEAParams}. We employ the gain design criterion proposed in Section~\ref{sec:gainselection} to design controller gains. Detailed stiffness and damping gains are accessible in Table~\ref{table:CriticalDampedGainSet}. All of our tests have a $1$ kHz sampling rate, which induces $0.5$ ms effective feedback delay. To obtain larger feedback delays, a software buffering of sampling data is manually implemented. Thus, the total feedback delay has two components
\begin{equation}
T_d = \frac{T_s}{2} + T_e,
\end{equation}
where $T_s$ is the sampling period and $T_e$ is the extra added feedback delay. $T_s$ is divided by 2 since the effective delay is half of the sampling period \cite{hulin2008stability}. The extra feedback delay, $T_e$, represents large round-trip communication delay between low-level and high-level architectures. The source code is public online \url{https://github.com/YeZhao/series-elastic-actuation-impedance-control}. Here is a video link of experimental validations \url{https://youtu.be/biIdlcAMPyE}.

\begin{figure}
 \centering
   \includegraphics[width=0.4\linewidth]{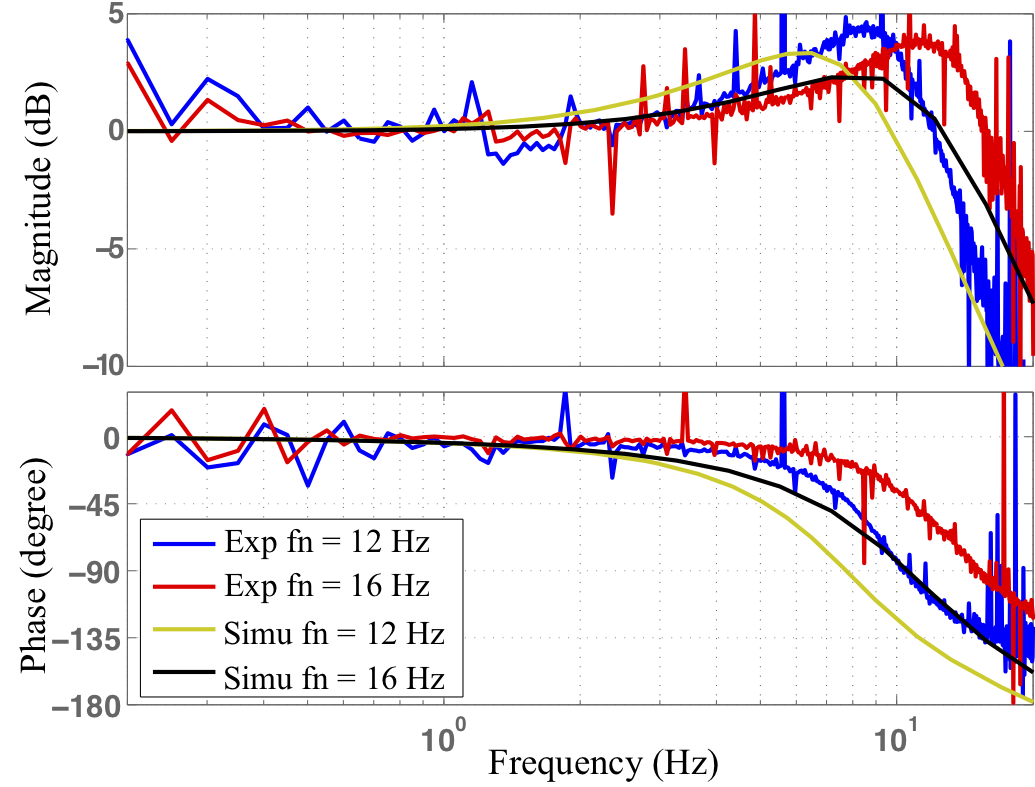}
\caption{\captionsize  Impedance frequency responses with different $f_n$. At low frequencies, experimental results are matched with the simulations. Compared to simulations, the experimental data shows a larger peak at the resonant frequency and a slightly larger bandwidth. The parameters are $T_{qs} = T_{qd} = T_{\tau} = 0.5$ ms, $f_{qd} = 50$ Hz, $f_{\tau d} = 100$ Hz, and $GS = 1$.}
 \label{fig:ExperimentBodeDiffFn}
\end{figure}

\begin{figure}[t]
 \centering
   \includegraphics[width=0.5\linewidth]{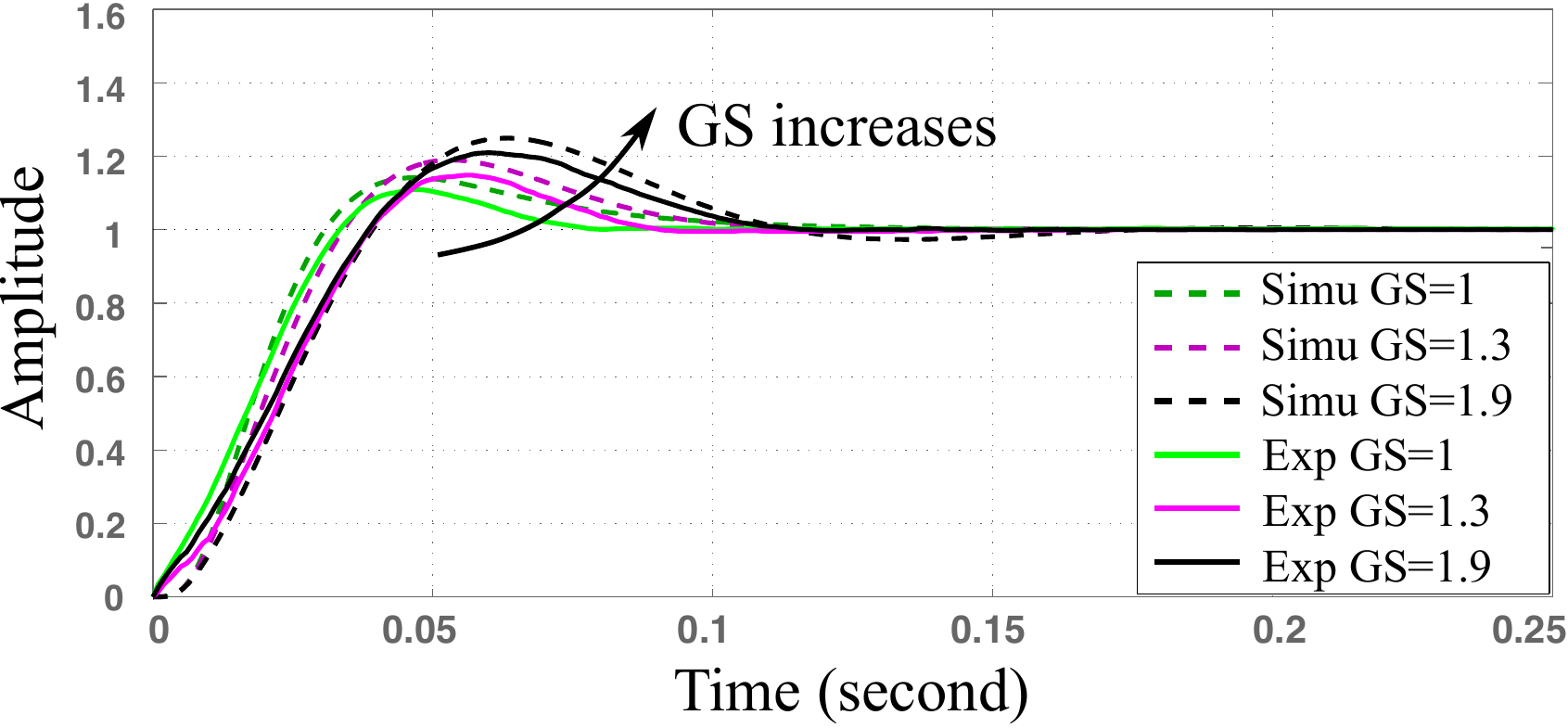}
\caption{\captionsize Step responses with different gain scales. The overshoot in the experimental results, when GS is increased, matches our simulation predictions. The parameters are $T_{qs} = T_{qd} = T_{\tau} = 1$ ms, $f_{qd} = 50$ Hz, $f_{\tau d} = 100$ Hz, and $f_n = 14$ Hz.
}
 \label{fig:ExperimentGS}
\end{figure}

In Fig.~\ref{fig:ExperimentBodeDiffFn}, a larger natural frequency produces a higher closed-loop bandwidth. Simulations match experimental results except slight discrepancies at high frequencies. To validate the trade-off between impedance gains and torque gains, we test step responses as shown in Fig.~\ref{fig:ExperimentGS}. The result shows that when $GS > 1$, a larger $GS$ slows down the rise time and produces a larger overshoot. This observation is consistent with our theoretical analysis that SEA phase margin will be reduced by decreasing impedance gains and increasing torque gains. As for the discrepancy between simulations and experiments, a potential reason is due to the different spring location in the simulation model and the hardware. The simulation model assumes the spring to be placed between the gearbox output and the load (a.k.a., force sensing SEA) while our UT-SEA hardware places the spring between the motor housing and the chassis ground (a.k.a., reaction force sensing SEA) for compact size design. This discrepancy affects impedance characteristics only at the resonant frequency and high-frequency, which is also validated by the result in Fig.~\ref{fig:ExperimentBodeDiffFn}. The reason why we choose a force sensing SEA model is due to being more general in the SEA literature, and more suitable for force control, and simplicity in the force measurements. For more details regarding these two mechanical designs, refer to \cite{painedesign2014}. The discrepancy between the two models is negligible in our tests since our primary target is to validate the trade-off between impedance and torque control.

\begin{figure}[t]
 \centering
   \includegraphics[width=0.7\linewidth]{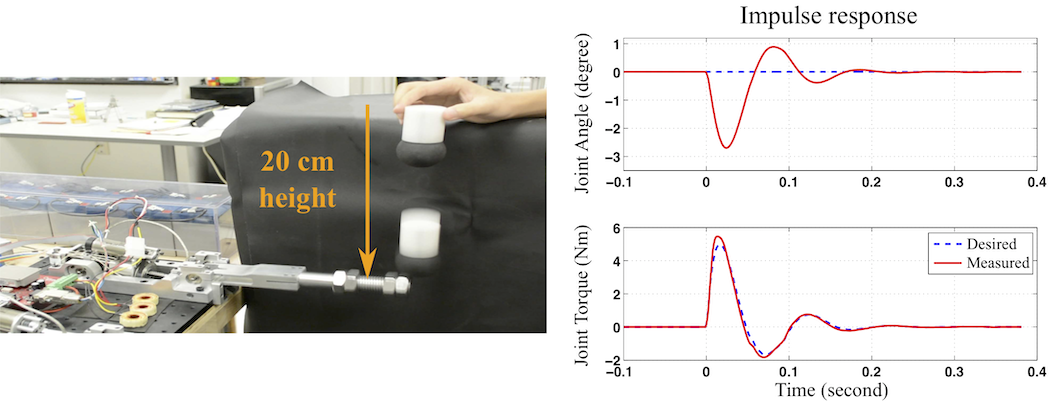}
\caption[Impulse response of UT-SEA]{\captionsize Impulse response of UT-SEA. A ball is dropped from a constant height (20 cm) and exerts an impulse force on the arm end-effector. The maximum angle deviation is around 2.5 degrees. The arm recovers to its initial position within 0.3 seconds. Joint torque tracking is accurate.}
 \label{fig:ExperimentImpulse}
 \vspace{-0.05in}
\end{figure}

Torque tracking under impact dynamics is important for interactive manipulation and bipedal locomotion. By implementing an impulse test, we show the high-fidelity of our torque control under external impulse disturbances. The purpose of this test is to performance of the controller under disturbances. The controller gains correspond to those of $f_n = 14$ Hz in Table~\ref{table:CriticalDampedGainSet}. As shown in Fig.~\ref{fig:ExperimentImpulse}, when a ball free falls from a 20 cm height and hits the arm with an impulse force, the SEA actuator settles down promptly and recovers after approximately 0.3 seconds. The recovery to the disturbance is fast and the tracking performance of the torque controller is very accurate.

\begin{table}[t]
\caption{UT SEA Parameters}
\centering
\vspace{0.1in}
\begin{tabular}{c|c||c|c}
\hline
Parameter & Value & Parameter & Value \\ \hline\hline
spring stiffness $k$ & $350000$ N/m & joint pulley radius $r_{k}$ & $0.025$ m \\  \hline 
motor inertia $I_m$ & $0.225$ kg$\cdot$m$^2$ & joint inertia $I_j$ & $0.014$ kg$\cdot$m$^2$ \\ \hline
motor damping $b_m$ & $1.375$ Nms/rad & joint damping $b_j$ & $0.1$ Nms/rad \\ \hline
gear reduction $N$ & $8.3776\times 10^{3}$  & ball screw lead $l_{bs}$ & 0.003 m/rev  \\  \hline 
drivetrain efficiency $\eta$ & $0.9$ & motor torque coeff $k_\tau$ & $0.0276$ Nm/A \\  \hline 
pulley reduction $N_p$ & 4 & sample rate & 1 kHz \\  \hline 
\end{tabular}
\label{table:SEAParams}
\end{table}

In the next subsection we study in detail the implementation of the proposed distributed control strategy in a high performance linear actuator and an omnidirectional mobile base.

\begin{figure}[t]
 \centering
   \includegraphics[width=0.5\linewidth]{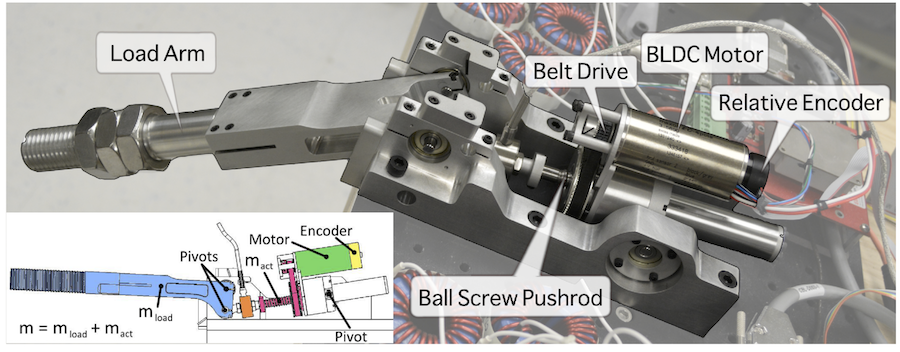}
 \caption{\captionsize Linear UT actuator. This linear pushrod actuator has an effective output inertia of $m = 256$ kg and an approximate passive damping of $b = 1250$ Ns/m.}
 \label{fig:UTActuator}
\end{figure}

\begin{figure}[t]
 \centering
   \includegraphics[width=0.6\linewidth]{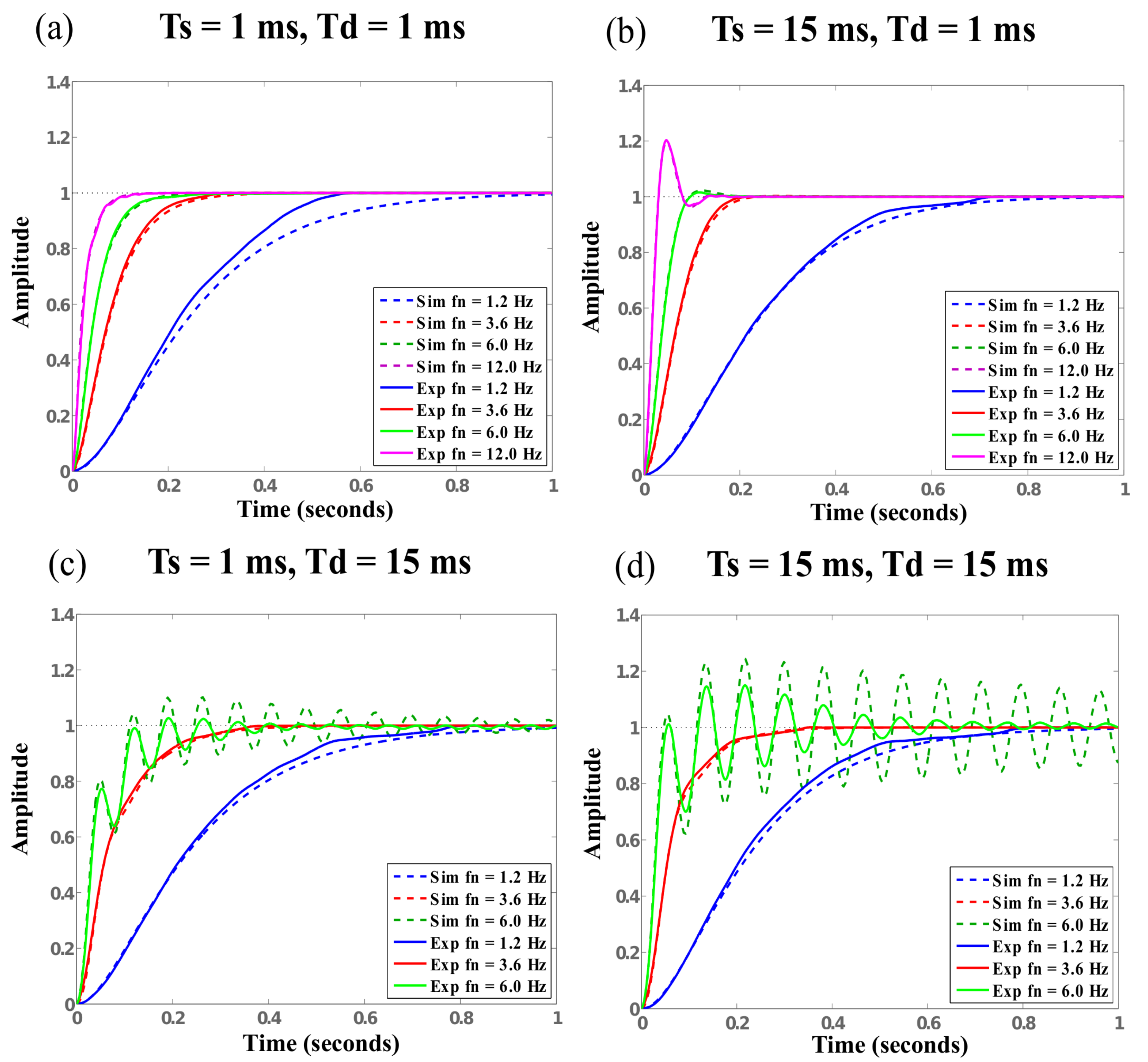}
 \caption{\captionsize Step response experiment with distributed controller. Subfigures (a) through (d) show various implementations on our linear actuator. Overlapped with the data plots, simulated replicas of the experiments are also shown to validate the proposed models. The experiments not only confirm the higher sensitivity of the actuator to damping than to stiffness delays but also indicate a good correlation between the real actuator and the simulations.}
 \label{fig:StepExpSimu_withoutweight}
\end{figure}

\begin{figure}[t]
 \centering
    \includegraphics[width=0.8\linewidth]{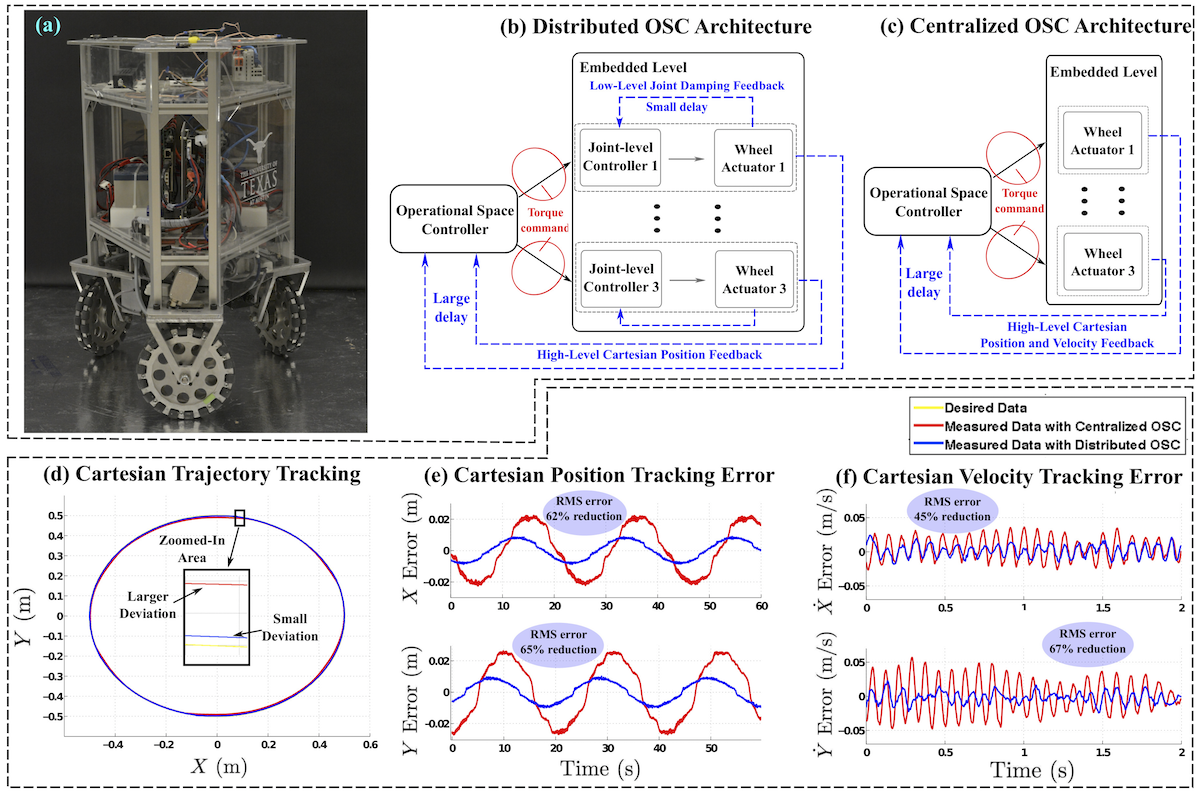}
 \caption{\captionsize Omnidirectional mobile base with distributed and centralized OSC controllers. As a proof of concept we leverage the proposed distributed architectures to our robotic mobile base demonstrating significant improvements on tracking and stability. }
 \label{fig:Trikey}
\end{figure}

\subsection{Step response implementation}
The proposed controller is implemented in our linear actuator shown in Figure~\ref{fig:UTActuator}. This actuator is equipped with a PC-104 form factor computer running Ubuntu Linux with an RTAI patched kernel \cite{painedesign2014}. The PC communicates with the actuator using analog and quadrature signals through a custom signal conditioning board. Continuous signal time derivatives are converted to discrete form using a bilinear Tustin transform written in C. A load arm is connected to the output of the ball screw pushrod. Small displacements enable the actuator to operate in an approximately linear region of its load inertia. At the same time, the controller is simulated by using the closed loop plant. Identical parameters to the real actuator are used for the simulation, thus allowing us to compare both side by side.

First, a test is performed on the actuator evaluating the response to a step input on its position. The results are shown in the bottom part of Figure~\ref{fig:StepExpSimu_withoutweight} which shows and compares the performance of the real actuator versus the simulated closed loop controller. 
All the experimental tests are performed with a $1$ kHz servo rate. Additional feedback delays are manually added by using a data buffer. A step input comprising desired displacements between $0.131$ m and $0.135$ m of physical pushrod length is sent to the actuator. The main reason for constraining the experiment to a small displacement is to prevent current saturation of the motor driver. With very high stiffness, it is easy to reach the $30$ A limit for step responses. If current is saturated, then the experiment will deviate from the simulation. The step response is normalized between $0$ and $1$ for simplicity. Various tests are performed for the same reference input with varying time delays. In particular large and small delays are used for either or both the stiffness and damping loops. The four combinations of results are shown in the figure with delay values of $1$ ms or $15$ ms.

The first thing to notice is that there is a good correlation between the real and the simulated results both for smooth and oscillatory behaviors. Small discrepancies are attributed to unmodelled static friction and the effect of unmodelled dynamics. More importantly, the experiment confirms the anticipated discrepancy in delay sensitivity between the stiffness and damping loops. Large servo delays on the stiffness servo, corresponding to subfigures (a) and (b) have small effects on the step response. On the other hand, large servo delays on the damping servo, corresponding to subfigures (c) and (d), strongly affect the stability of the controller. In fact, for (c) and (d) the results corresponding to $f_n = 12$ Hz are omitted due to the actuator quickly becoming out of control. In contrast, the experiment in (b) can tolerate such high gains despite the large stiffness delay. 

\subsection{Distributed operational space control of a mobile base}
\label{subsec:doscmb}
As a concept proof of the proposed distributed architecture on a multi-axis mobile platform, a Cartesian space feedback Operational Space Controller \cite{Khatib:87(2)} is implemented on an omnidirectional mobile base. The original feedback controller was implemented as a centralized process with no distributed topology at that time. The mobile base is equipped with a centralized PC computer running Linux with the RTAI real-time kernel. The PC connects with three actuator processors embedded next to the wheel drivetrains via EtherCat serial communications. The embedded processors do not talk to each other. The high level centralized PC on our robot, has a roundtrip latency to the actuators of 7ms due to process and bus communications, while the low level embedded processors have a servo rate of 0.5ms. Notice that 7ms is considered too slow for stiff feedback control. To accentuate even further the effect of feedback delay on the centralized PC, an additional 15ms delay is artificially introduced by using a data buffer. Thus, the high level controller has a total of $22$ ms feedback delay.

\begin{figure*}
\centering
\includegraphics[width = 0.8\linewidth]{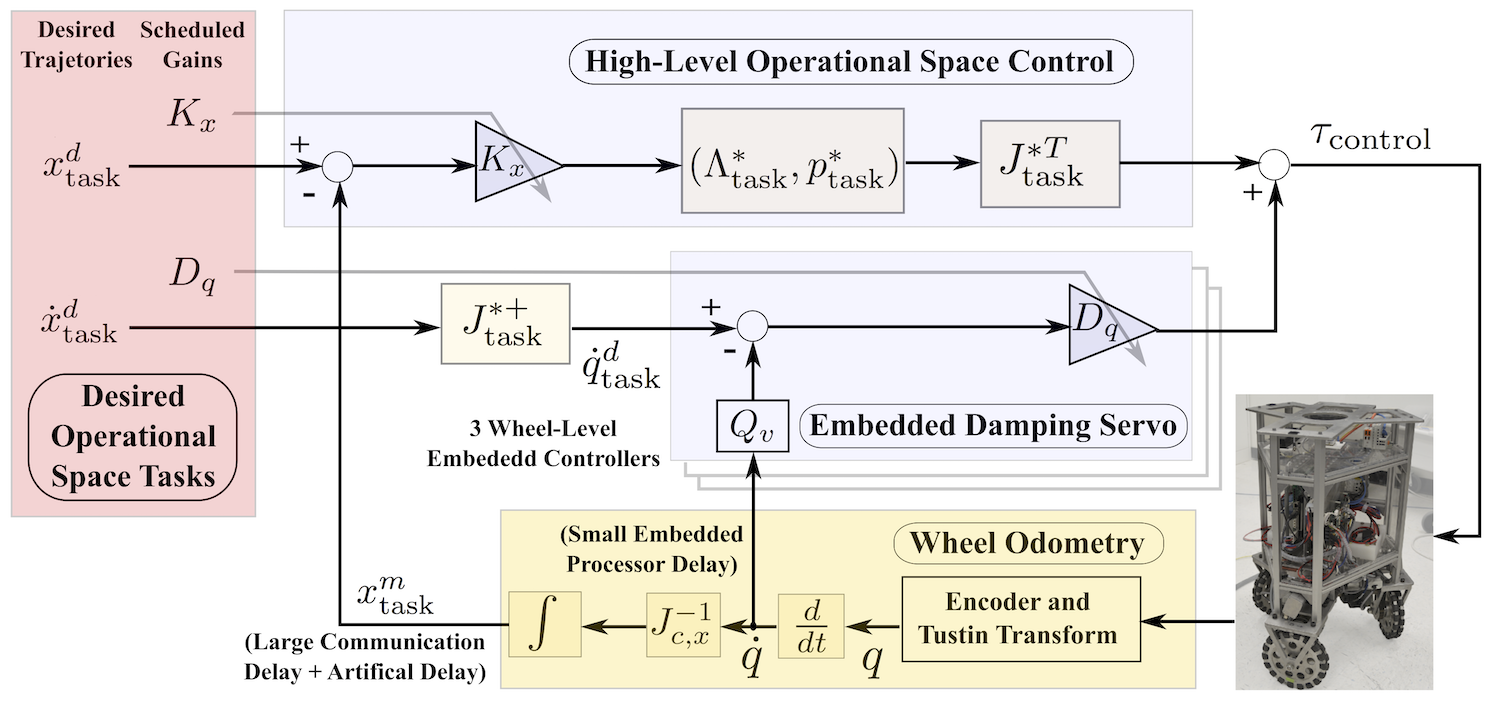}
\caption{\captionsize Detailed distributed operational space control structure. The figure above illustrates details of the distributed operational space controller used for the mobile base tracking experiment. $\Lambda^*_{\rm task}$ and $p^*_{\rm task}$ are the operational space inertia matrix and gravity based forces, respectively. $J^{*}_{\rm task}$ is a contact consistent task Jacobin. More details about these matrices and vectors can be found in \cite{Khatib:87(2)}. Our main contribution for this experiment lies in implementing operational space control in a distributed fashion and based on the observations performed on the previously simplified distributed controller. While the high-level operational space stiffness feedback loop suffers from large delays due to communication latencies and artificial delays (added by a data buffer), the embedded-level damping loop increases system stability. As a result, the proposed distributed architecture enables to achieve higher Cartesian stiffness gains $K_x$ for better tracking accuracy.}
\label{fig:DOSC}
\end{figure*}

An operational space controller (OSC) is implemented in the mobile base using two different architectures. First, the controller is implemented as a centralized process, which will be called COSC, with all feedback processes taking place in the slow centralized processor and none in the embedded processors. In this case, the maximum stiffness gains should be severely limited due to the effect of the large latencies. Second a distributed controller architecture is implemented inspired by the one proposed in Figure~\ref{fig:ControlBlock} but adapted to a desired operational space controller, which will be called DOSC. In this version, the Cartesian stiffness feedback servo is implemented in the centralized PC in the same way than in COSC, but the Cartesian damping feedback servo is removed from the centralized process. Instead, our study implements damping feedback in joint space (i.e. proportional to the wheel velocities) on the embedded processors. A conceptual drawing of these architectures is shown in Figure~\ref{fig:Trikey}. The metric used for performance comparison is based on the maximum achievable Cartesian stiffness feedback gains, and the Cartesian position and velocity tracking errors.

To implement the Cartesian stiffness feedback processes in both architectures, the Cartesian positions and orientations of the mobile base on the ground are computed using wheel odometry. To achieve the highest stable stiffness gains, the following procedure is followed: (1) first, Cartesian stiffness gains are adjusted to zero while the damping gains in either Cartesian space (COSC) or joint space (DOSC) -- depending on the controller architecture -- are increased until the base starts vibrating; (2) the Cartesian stiffness gains, on either architecture, are increased until the base starts vibrating or oscillating; (3) a desired Cartesian circular trajectory is commanded to the base and the position and velocity tracking performance are recorded.

Based on these experiments, DOSC was able to attain a maximum Cartesian stiffness gain of 140$\,$N/(m $\,$kg) compared to 30$\,$N/(m$\,$kg) for COSC. This result means that the proposed distributed control architecture allowed the Cartesian feedback process to increase the Cartesian stiffness gain ($K_x$ in Figure~\ref{fig:DOSC}) by 4.7 times with respect to the centralized controller implementation. In terms of tracking performance, the results are shown in Figure~{\ref{fig:Trikey}}. Both Cartesian position and velocity tracking in DOSC are significantly more accurate. The proposed distributed architecture reduces Cartesian position root mean error between 62\% and 65\% while the Cartesian velocity root mean error decreases between 45\% and 67\%.

\section{Discussions and Conclusion}
The motivation for this chapter has been to study the stability and performance of distributed controllers where stiffness and damping servos are implemented in distinct processors. These types of controllers will become important as computation and communications become increasingly more complex in human-centered robotic systems. The focus has been first on studying the physical performance of a simple distributed controller. Simplifying the controller allows us to explore the physical effects of time delays in greater detail. Based on this controller, we address the problem of impedance controller design and performance characterization of series elastic actuators (SEAs) by incorporating time delays and filtering over a wide frequency spectrum. In particular, we proposed a critically-damped controller gain selection method of the cascaded SEA control structure. By uncovering the trade-off existing between impedance gains and torque gains, we prove the optimality of our gain design criterion. We believe the critically-damped gain selection criterion can be applied to many types of SEAs and robotics systems for performance analysis and optimizations. 

To confirm the observations and analytical derivations, hardware experiments are performed by using an actuator and a mobile base. In particular, the results have shown that decoupling stiffness servos to slower centralized processes does not significantly decrease system stability. As such, stiffness servo can be used to implement operational space controllers which require centralized information such as robot models and external sensors.

\bibliographystyle{IEEEtran}
\bibliography{bib}

\end{document}